\documentclass{article}

\usepackage{microtype}
\usepackage{graphicx}
\usepackage{subfigure}
\usepackage{booktabs} % for professional tables

\usepackage{hyperref}

\usepackage[ruled,vlined]{algorithm2e}
\usepackage{listings}

\usepackage{subcaption}
\usepackage[font=small,labelfont=bf]{caption}
\usepackage{booktabs}

\usepackage[accepted]{icml2024}

\usepackage{amsmath}
\usepackage{amssymb}
\usepackage{mathtools}
\usepackage{amsthm}
\usepackage{physics}

\usepackage[capitalize,noabbrev]{cleveref}

\theoremstyle{plain}

\theoremstyle{definition}

\theoremstyle{remark}

\DeclareMathOperator*{\argmin}{arg\,min}

\usepackage[textsize=tiny]{todonotes}

\icmltitlerunning{Variance-Covariance Regularization}

\begin{document}

\twocolumn[
\icmltitle{Variance-Covariance Regularization Improves Representation Learning}


\begin{icmlauthorlist}
\icmlauthor{Jiachen Zhu}{nyu-cs}
\icmlauthor{Katrina Evtimova}{nyu-cds}
\icmlauthor{Yubei Chen}{ucd}
\icmlauthor{Ravid Shwartz-Ziv}{nyu-cds}
\icmlauthor{Yann LeCun}{nyu-cs,nyu-cds,meta}
\end{icmlauthorlist}

\icmlaffiliation{nyu-cs}{New York University, Computer Science Department}
\icmlaffiliation{nyu-cds}{New York University, Center for Data Science}
\icmlaffiliation{meta}{Meta AI, FAIR}
\icmlaffiliation{ucd}{UC Davis, Electrical and Computer Engineering Department}

\icmlcorrespondingauthor{Jiachen Zhu}{jiachen.zhu@nyu.edu}

\icmlkeywords{Machine Learning, ICML, Normalization}

\vskip 0.3in
]

\printAffiliationsAndNotice

\begin{abstract}
Transfer learning plays a key role in advancing machine learning models, yet conventional supervised pretraining often undermines feature transferability by prioritizing features that minimize the pretraining loss. In this work, we adapt a self-supervised learning regularization technique from the VICReg method to supervised learning contexts, introducing Variance-Covariance Regularization (VCReg). This adaptation encourages the network to learn  high-variance, low-covariance representations, promoting learning more diverse features. We outline best practices for an efficient implementation of our framework, including applying it to the intermediate representations. Through extensive empirical evaluation, we demonstrate that our method significantly enhances transfer learning for images and videos, achieving state-of-the-art performance across numerous tasks and datasets. VCReg also improves performance in scenarios like long-tail learning and hierarchical classification. Additionally, we show its effectiveness may stem from its success in addressing challenges like gradient starvation and neural collapse. In summary, VCReg offers a universally applicable regularization framework that significantly advances transfer learning and highlights the connection between gradient starvation, neural collapse, and feature transferability.
\end{abstract}

\section{Introduction}

\begin{figure}
  \centering
  \includegraphics[height=40mm, width=82mm]{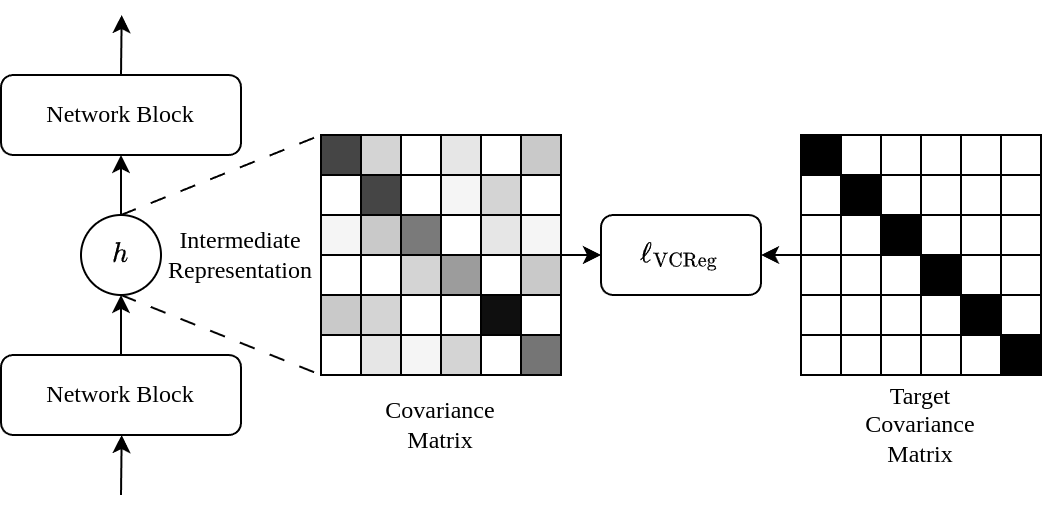}
  \caption{\textbf{VCReg regularizes the network by encouraging the intermediate representations to have high variance and low covariance.} VCReg is applied to the output of each network block to make all the intermediate representations capture diverse features.}
  \label{vcr}
\end{figure}

Transfer learning enables models to apply knowledge from one domain to enhance performance in another, particularly when data are scarce or costly to obtain \citep{pan2010survey, Weiss2016ASO, zhuang2020comprehensive, bommasani2021opportunities}. One of the key challenges arises during the supervised pretraining phase. In this phase, models often lack detailed information about the downstream tasks to which they will be applied. Nevertheless, they must aim to capture a broad spectrum of features beneficial across various applications \citep{bengio2012deep, caruana1997multitask, yosinski2014transferable}. Without proper regularization techniques, these supervised pretrained models tend to overly focus on features that minimize supervised loss, resulting in limited generalization capabilities and issues such as gradient starvation and neural collapse \citep{zhang2016understanding, neyshabur2017exploring, zhang2021understanding, pezeshki2021gradient, papyan2020prevalence, shwartz2022information}.

To tackle these challenges, we adapt the regularization techniques of the self-supervised VICReg method \citep{bardes2021vicreg} for the supervised learning paradigm. Our method, termed Variance-Covariance Regularization (VCReg), aims to encourage the learning of representations with high variance and low covariance, thus avoiding the overemphasis on features that merely minimize supervised loss. Instead of simply applying VCReg to the final representation of the network, we explore the most effective ways to incorporate it throughout the intermediate representations.

The structure of the paper is as follows: we begin with an introduction of our method, including an outline of a fast implementation strategy designed to minimize computational overhead. Following this, we present a series of experiments aimed at validating the method's efficacy across a wide range of tasks, datasets, and architectures. Subsequently, we conduct analyses on the learned representations to demonstrate VCReg's effectiveness in mitigating common issues in transfer learning, such as neural collapse and gradient starvation. 

Our paper makes the following contributions:
\begin{enumerate}
\item We introduce a robust strategy for applying VCReg to neural networks, including integrating it into the intermediate layers.
\item We propose a computationally efficient implementation of VCReg. This implementation is optimized to ensure minimal additional computational overhead, allowing for seamless integration into existing workflows.
\item Through extensive experiments on benchmark datasets both in images and videos, we demonstrate that VCReg suppresses the prior state-of-the-art results in transfer learning performance across various network architectures, including ResNet \citep{he2016deep}, ConvNeXt \citep{liu2022convnet}, and ViT \citep{dosovitskiy2020image}. Moreover, we also show that VCReg improves performance in diverse scenarios like long-tail learning and hierarchical classification.
\item We investigate the representations learned by VCReg, revealing its effectiveness in combating challenges such as gradient starvation \citep{pezeshki2021gradient}, neural collapse \citep{papyan2020prevalence}, information compression \citep{shwartz2022information}, and sensitivity to noise.
\end{enumerate}

Before delving into VCReg's details in the following sections, it is key to note its divergence from VICReg, namely by omitting the invariance loss and focusing on variance and covariance loss for a wider application, especially in transfer learning. This approach tackles challenges like gradient starvation and neural collapse, advancing neural network training across various architectures. Our work further distinguishes itself by exploring optimal regularization strategies, moving beyond generic application to significantly enhance its 
effectiveness.

\section{Related Work}

\subsection{Variance-Invariance-Covariance Regularization (VICReg)}
VICReg \citep{bardes2021vicreg} is a novel SSL method that encourages the learned representation to be invariant to data augmentation. However, focusing solely on this invariance criterion can result in the network producing a constant representation, making it invariant to both data augmentation and the input data itself.

VICReg primarily regularizes the network by combining variance loss and covariance loss. The variance loss encourages high variance in the learned representations, thereby promoting the learning of diverse features. The covariance loss, on the other hand, aims to minimize redundancy in the learned features by reducing the overlap in information captured by different dimensions of the representation. This dual-objective optimization framework effectively promotes diverse feature learning for SSL \citep{shwartz2022we}. To improve the performance of supervised network training, we adapt the SSL feature collapse prevention mechanism from VICReg and propose a variance-covariance regularization method.

To calculate the loss function of VICReg with a batch of data \(\{x_1 \ldots x_n\}\), we first need to have a pair of inputs \((x'_i, x''_i)\) such that \(x'_i\) and \(x''_i\) are two augmented versions of the original input \(x_i\). Given the neural network \(f_\theta(\cdot)\) and the final representations \(z'_i = f_\theta(x'_i)\) and \(z''_i = f_\theta(x''_i)\) such that $z'_i, z''_i \in \mathbb{R}^D$, VICReg minimizes the following loss:
\begin{align}
&\ell_{\mathrm{VICReg}}(z'_1 \ldots z'_n, z''_1 \ldots z''_n) \\
&= \alpha \ell_{\mathrm{var}}(z'_1, \ldots, z'_n) + \alpha \ell_{\mathrm{var}}(z''_1, \ldots, z''_n) \label{eq:vicreg}\\ 
&+ \beta \ell_{\mathrm{cov}}(z'_1, \ldots, z'_n) + \beta \ell_{\mathrm{cov}}(z''_1, \ldots, z''_n) \nonumber \\ 
&+ \sum_{i=1}^n \ell_{\mathrm{inv}}(z'_i, z''_i). \nonumber
\end{align}
The variance and covariance loss functions are defined as:
\begin{align}
\ell_{\mathrm{var}} &= \frac{1}{D} \sum_{i=1}^{D} \max(0, 1 - \sqrt{C_{ii}}) \label{eq:var}\\
\ell_{\mathrm{cov}} &= \frac{1}{D(D-1)} \sum_{i \neq j} C_{ij}^2 \label{eq:cov}
\end{align}
where \(C = \frac{1}{N-1}\sum_{i=1}^N (z_i - \bar{z})(z_i - \bar{z})^T\) denotes the covariance matrix, and \(\bar{z}\) represents the mean vector, given by \(\bar{z} = \frac{1}{N} \sum_{i=1}^N z_i\). 

Building on insights from prior studies \citep{shwartz2022information, shwartz2023information}, it is understood that the invariance term does not play a pivotal role in diversifying features. Consequently, in adapting to the supervised regime, we exclude the invariance term from the regularization.

\subsection{Representation Whitening and Feature Diversity Regularizers}

Representation whitening is a technique for processing inputs before they enter a network layer. It transforms the input so that its components are uncorrelated with unit variance \citep{kessy2018optimal}. This transformation achieves enhanced model optimization and generalization. It uses a whitening matrix derived from the data's covariance matrix and results in an identity covariance matrix, thereby aiding gradient flow during training and acting as a lightweight regularizer to reduce overfitting and encourage robust data representations \citep{lecun2002efficient}.

In addition to whitening as a processing step, additional regularization terms can be introduced to enforce decorrelation in the representations. Various prior works have explored these feature diversity regularization techniques to enhance neural network training \citep{cogswell2015reducing, ayinde2019regularizing, laakom2023wld}. These methods encourage diverse features in the representation by adding a regularization term. Recent methods like WLD-Reg \citep{laakom2023wld} and DeCov \citep{cogswell2015reducing} also employ covariance-matrix-based regularization to promote feature diversity, similarly to our approach.

However, the studies above mainly focus on the benefits of optimization and generalization for the source task, often neglecting their implications for supervised transfer learning. VCReg distinguishes itself by explicitly targeting enhancements in transfer learning performance. Our results indicate that such regularization techniques yield only modest performance improvements in in-domain evaluations.

\subsection{Gradient Starvation and Neural Collapse}

Gradient starvation and neural collapse are two recently recognized phenomena that can significantly affect the quality of learned representations and a network's generalization ability \citep{pezeshki2021gradient, papyan2020prevalence, ben2023reverse}. 
Gradient starvation occurs when certain parameters in a deep learning model receive very small gradients during the training process, thereby leading to slower or non-existent learning for these parameters \citep{pezeshki2021gradient}. 
Neural collapse, on the other hand, is a phenomenon observed during the late stages of training when the internal representations of the network tend to collapse towards each other, resulting in a loss of feature diversity \citep{papyan2020prevalence}. 
Both phenomena are particularly relevant in the context of transfer learning, where models are initially trained on a source task before being fine-tuned for a target task. Our work, through the use of VCReg, seeks to mitigate these issues, offering a pathway to more effective transfer learning.

\section{Variance-Covariance Regularization}

\subsection{Vanilla VCReg}

Consider a labeled dataset comprising \(N\) samples, denoted as \(\{(x_1, y_1)\ldots (x_N, y_N)\}\) and a neural network \(f_\theta(\cdot)\), which takes these inputs \(x_i\) and produces final predictions \(\tilde{y}_i = f_\theta(x_i)\). In standard supervised learning, the loss is defined as $L_{\mathrm{sup}} = \frac{1}{N}\sum_{i=1}^N \ell_{\mathrm{sup}}(\tilde{y}_i, y_i)$.

The core objective of the Vanilla VCReg is to ensure that the \(D\)-dimensional input representations \(\{h_i\}_{i=1}^N\) to the last layer of the network exhibit both high variance and low covariance. To achieve this, we employ variance and covariance regularization, same as mentioned in equation \ref{eq:vicreg}:
\begin{align}
\ell_{\mathrm{vcreg}}(h_1 \ldots h_N) &= \alpha \ell_{\mathrm{var}}(h_1 \ldots h_N) + \beta \ell_{\mathrm{cov}}(h_1 \ldots h_N)
\end{align}

Intuitively speaking, the covariance matrix captures the interdependencies among the dimensions of the feature vectors \(h_i\). Maximizing \(\ell_{\mathrm{var}}\) encourages each feature dimension to contain unique, non-redundant information, while minimizing \(\ell_{\mathrm{cov}}\) aims to reduce the correlation between different dimensions, thus promoting feature independence. The overall training loss, which includes also the supervised loss, then becomes:
\begin{align}
L_{\mathrm{vanilla}} &= \alpha \ell_{\mathrm{var}}(h_1\ldots h_N) + \beta \ell_{\mathrm{cov}}(h_1\ldots h_N) \\
&+ \frac{1}{N} \sum_{i=1}^N \ell_{\mathrm{sup}}(\tilde{y}_i, y_i).
\end{align}
Here, \(\alpha\) and \(\beta\) serve as hyperparameters to control the strength of each regularization term.

\subsection{Extending VCReg to Intermediate Representations}

While regularizing the final layer in a neural network offers certain benefits, extending this approach to intermediate layers via VCReg provides additional advantages (for empirical evidence supporting this claim, please refer to Appendix \ref{investigation}). Regularizing intermediate layers enables the model to capture more complex, higher-level abstractions. This strategy minimizes internal covariate shifts across layers, which in turn improves both the stability of training and the model's generalization capabilities. Furthermore, it fosters the development of feature hierarchies and enriches the latent space, leading to enhanced model interpretability and improved transfer learning performance.

To implement this extension, VCReg is applied at \( M \) strategically chosen layers throughout the neural network. For each intermediate layer \( j \), we denote the feature representation for an input \( x_i \) as \( h_i^{(j)} \in \mathbb{R}^{D_j} \). This culminates in a composite loss function, expressed as follows:
\begin{align}
L_{\text{VCReg}} &= \sum_{j=1}^M \left [ \alpha \ell_{\text{var}}(h_1^{(j)} \ldots h_N^{(j)}) + \beta \ell_{\text{cov}}(h_1^{(j)} \ldots h_N^{(j)}) \right ] \\
&+ \frac{1}{N}\sum_{i=1}^N \ell_{\text{sup}}(\tilde{y}_i, y_i).
\end{align}

\textbf{Spatial Dimensions} However, applying VCReg to intermediate layers of real-world neural networks presents challenges due to the spatial dimensions in these intermediate representations. Naively reshaping these representations into long vectors would lead to unmanageably large covariance matrices, thereby increasing computational costs and risking numerical instability.
To address this issue, we adapt VCReg to accommodate networks with spatial dimensions. Each vector at a different spatial location is treated as an individual sample when calculating the covariance matrix. Both the variance loss and the covariance loss are then calculated based on this modified covariance matrix.

In terms of practical implementation, a VCReg is usually applied subsequently to each block within the neural network architecture, often succeeding residual connections. This placement allows for seamless incorporation into current network architectures and training paradigms.

\textbf{Addressing Outliers with Smooth L1 Loss} After treating spatial locations as independent samples for covariance computation, the resulting samples are no longer statistically independent. This can lead to outliers in the covariance matrix and unstable gradient updates. To address this, we introduce a smooth L1 penalty into the covariance loss term. Specifically, we replace the traditional squared covariance values \( C_{ij} \) in \( \ell_{\text{cov}} \) with a smooth L1 function:
\begin{align}
\text{SmoothL1}(x) = 
\begin{cases}
    x^2, & \text{if } |x| \leq \delta \\
    2 \delta |x| - \delta^2, & \text{otherwise}
\end{cases}
\end{align}

By implementing this modification, we ensure that the loss function increases in a more controlled manner with respect to large covariance values. Empirically, this minimizes the impact of outliers, thereby enhancing the stability of the training process.

\subsection{Fast Implementation}

To optimize VCRef speed, we use the fact that VCReg only affects the loss function and not the forward pass. This allows us to focus on directly modifying the backward function for improvements. Specifically, we sidestep the usual process of calculating the VCReg loss and subsequent backpropagation. Instead, we directly adjust the computed gradients, which is feasible since the VCReg loss calculation relies solely on the current representation. Further details of this speed-optimized technique are outlined in Appendix \ref{sec:appendix-fast}. Our optimized VCReg implementation exhibits similar latency as batch normalization layers and is more than 5 times faster than the naive VCReg implementation. The results are presented in Table \ref{ta:time}.

\section{Experiments}

In this section, we first outline the experimental framework and findings highlighting the effectiveness of our proposed regularization approach, VCReg, within the realm of transfer learning that utilizes supervised pretraining for both images and videos. Subsequently, we extend our experiments to three specialized learning scenarios: 1) class imbalance via long-tail learning, 2) synergizing with self-supervised learning frameworks, and 3) hierarchical classification problems. The objective is to assess the adaptability of VCReg across various data distributions and learning paradigms, thereby evaluating its broader utility in machine learning applications. For details on reproducing our experiments, please consult Appendix \ref{sec:appendix-detail}.

\subsection{Transfer Learning for Images}

In this section, we adhere to evaluation protocols established by seminal works such as \citep{chen2020simple, kornblith2021better, misra2020self} for our transfer learning experiments.

Initially, we pretrain models using three different architectures: ResNet-50 \citep{he2016deep}, ConvNeXt-Tiny \citep{liu2022convnet}, and ViT-Base-32 \citep{dosovitskiy2020image}, on the full ImageNet dataset. We follow the standard PyTorch recipes \citep{NEURIPS2019_9015} for all networks and do not modify any hyperparameters other than those related to VCReg to ensure a fair baseline comparison.
Subsequently, we perform a linear probing evaluation across 9 different benchmark  to evaluate the transfer learning performance.

For ResNet-50, we include two other feature diversity regularizer methods for comparison: DeCov \citep{cogswell2015reducing} and WLD-Reg \citep{laakom2023wld}. We conduct experiments solely with ResNet-50 because it is the principal architecture used in the WLD-Reg paper. To ensure a fair comparison, we source hyperparameters from \citet{laakom2023wld} for both DeCov and WLD-Reg.

\begin{table*}[t]
  \caption{\textbf{Transfer Learning Performance with ImageNet Supervised Pretraining} The table shows performance metrics for different architectures. Each model is pretrained on the full ImageNet dataset and then tested on different downstream datasets using linear probing. Application of VCReg consistently improves performance and beats other feature diversity regularizer. Averages are calculated excluding ImageNet results.}
  \label{transfer_learning_1}
  \vskip 0.1in
  \centering
  \scalebox{0.9}{
  \begin{tabular}{lllllllllll}
    \toprule
    Architecture & iNat18 & Places & Food & Cars & Aircraft & Pets & Flowers & DTD & {Average} \\
    \midrule
    ResNet-50 & 42.8\% & 50.6\% & 69.1\% & 43.6\% & 54.8\% & 91.9\% & 77.1\% & 68.7\% & 62.33\% \\
    ResNet-50 (DeCov) & 43.1\% & 50.4\% & 69.0\% & 45.7\% & 55.5\% & 90.6\% & 79.2\% & 69.1\% & 62.83\% \\
    ResNet-50 (WLD-Reg) & 43.9\% & \textbf{51.2\%} & 70.2\% & 43.9\% & 58.7\% & 91.4\% & 80.7\% & 69.0\% & 63.63\% \\
    \textbf{ResNet-50 (VCReg)} & \textbf{45.3\%} & \textbf{51.2\%} & \textbf{71.7\%} & \textbf{54.1\%} & \textbf{70.5\%} & \textbf{92.1\%} & \textbf{88.0\%} & \textbf{70.8\%} & \textbf{67.96\%} \\
    \midrule
    ConvNeXt-T & 51.6\% & 53.8\% & 78.4\% & 62.9\% & 74.7\% & 93.9\% & 91.3\% & 72.9\% & 72.44\% \\
    \textbf{ConvNeXt-T (VCReg)} & \textbf{52.3\%} & \textbf{54.7\%} & \textbf{79.6\%} & \textbf{64.2\%} & \textbf{76.3\%} & \textbf{94.1\%} & \textbf{92.7\%} & \textbf{73.3\%} & \textbf{73.40\%} \\
    \midrule
    ViT-Base-32 & 39.1\% & 47.9\% & 70.6\% & 51.2\% & 63.8\% & 90.3\% & 84.6\% & 66.1\% & 64.20\% \\
    \textbf{ViT-Base-32 (VCReg)} & \textbf{40.6\%} & \textbf{48.1\%} & \textbf{70.9\%} & \textbf{52.0\%} & \textbf{65.8\%} & \textbf{91.0\%} & \textbf{86.6\%} & \textbf{66.5\%} & \textbf{65.19\%} \\
    \bottomrule
  \end{tabular}}
\end{table*}

The results in Table \ref{transfer_learning_1} demonstrate that VCReg significantly enhances performance in transfer learning for images across almost all downstream datasets, achieving the highest performance for 9 out of 10 datasets, and for all three architectures. Clearly, VCReg acts as a versatile plug-in, effectively boosting transfer learning outcomes. Its effectiveness spans ConvNet and Transformer architectures, confirming its wide-ranging applicability.

\subsection{Transfer Learning for Videos}
\label{sec:transfer-with-video}
To extend our evaluation of VCReg's efficacy, we conduct additional experiments using networks pretrained on video datasets. Specifically, we utilized models pretrained on Kinetics-400 \cite{kay2017kinetics} and Kinetics-710 \cite{li2022uniformerv2}, subsequently finetuning them for action recognition tasks on HMDB51 \cite{kuehne2011hmdb}. This set of experiments encompassed models trained with self-supervised learning objectives, including VideoMAE \cite{tong2022videomae} and VideoMAEv2 \cite{wang2023videomae}, as well as models trained with conventional supervised learning objectives, such as ViViT \cite{arnab2021vivit}.

We follow the finetuning protocols detailed by \citet{tong2022videomae} and the conventional evaluation method used in the field, where the final performance is measured by the mean classification accuracy across three provided splits \cite{simonyan2014two}. To pinpoint the optimal VCReg coefficients, we conducted a grid search based on validation set accuracy. For simplicity, in this setup, VCReg regularization is exclusively applied to the final output of each network during finetuning, just before the classification head.

Table \ref{tab:transfer_learning_video} illustrates that incorporating VCReg as a plugin regularizer enhances video classification performance across various methods (VideoMAE, VideoMAE2, and ViViT-B) and backbone architectures (ViT-B and ViT-S). The performance gains are evident in the improvements seen with VCReg across all models in the table. This consistent enhancement across a spectrum of models solidifies VCReg's status as a practical and versatile regularizer, capable of substantially improving the performance of pretrained networks in transfer learning scenarios.

% We apply VCReg only during fine-tuning to the final output of each network preceding the classification head. For both HMDB51, we follow the fine-tuning protocols outlined by \cite{tong2022videomae}. We determine the optimal values of the VCReg coefficients through grid search based on the validation set accuracy. We follow the standard evaluation protocol in the literature in which the final performance is measured by the mean classification accuracy across the three provided splits \cite{simonyan2014two}. Table \ref{tab:transfer_learning_video} shows that VCReg can boost classification performance when transfering features of networks pre-trained on video data.

\begin{table}[ht]
\caption{\textbf{Transfer Learning Performance with Kinetics-400 and Kinetics-710 pretrained models}: The table shows fine-tuning performance of Kinetics pre-trained models on HMDB51. VideoMAE-S, VideoMAE-B, and ViViT-B are pretrained on Kinetics-400 dataset while VideoMAEv2-S and VideoMAEv2-B are pre-trained on Kinetics-710. We apply VCReg only to the networks' output preceding the classification head. The results show that VCReg can boost the transfer learning classification performance for networks pre-trained on video data.}
\label{tab:transfer_learning_video}
\centering
\vskip 0.1in
\begin{small}
\begin{tabular}{llll}
\toprule
Method     & Backbone           & HMDB51  \\\midrule
VideoMAE-S           & ViT-S             & 79.9\%                      \\
VideoMAE-S (VCReg)       & ViT-S         & \textbf{80.6\%}                     \\\midrule
VideoMAE-B            & ViT-B             & 82.2\%                      \\
VideoMAE-B (VCReg)   & ViT-B            & \textbf{83.0\%}                     \\\midrule
VideoMAEv2-S        & ViT-S               & 83.6\%                     &  \\
VideoMAEv2-S (VCReg)  & ViT-S            & \textbf{83.9\%}                    \\\midrule
VideoMAEv2-B      & ViT-B           & 86.5\%                     &      \\
VideoMAEv2-B (VCReg)   & ViT-B           & \textbf{86.9\%}                    &      \\\midrule
ViViT-B         & ViT-B               & 70.9\%                     \\
ViViT-B (VCReg)    & ViT-B               & \textbf{71.6\%}                   \\
\bottomrule
\end{tabular}
\vskip -0.1in
\end{small}
\end{table}

\subsection{Class Imbalance with Long-Tail Learning}

Class imbalance is a pervasive issue in many real-world datasets and poses a considerable challenge to standard neural network training algorithms. We conduct experiments to assess how well VCReg addresses this issue through long-tail learning.
We evaluate VCReg using the CIFAR10-LT and CIFAR100-LT \cite{krizhevsky2009learning} datasets, both engineered to have an imbalance ratio of 100. These experiments use a ResNet-32 backbone architecture. The per-class sample sizes ranges from 5,000 to 50 for CIFAR10-LT and from 500 to 5 for CIFAR100-LT.

\begin{table}[h]
\label{table:lt}
%\vskip 0.1in
\vskip .3in

\begin{center}
\begin{small}
\begin{tabular}{lccr}
\toprule
Training Methods & CIFAR10-LT  & CIFAR100-LT \\
\midrule
ResNet-32  & 69.6\%  & 37.4\% \\
ResNet-32 (VCReg)  & \textbf{71.2\%}  & \textbf{40.4\%} \\
\bottomrule
\end{tabular}

\caption{\textbf{Performance Comparison on Class-Imbalanced Datasets Using VCReg}: This table shows the accuracy of standard ResNet-32 with and without VCReg when trained on class-imbalanced CIFAR10-LT and CIFAR100-LT datasets. The VCReg-enhanced models show improved performance, demonstrating the method's effectiveness in addressing class imbalance.}
\end{small}
\end{center}
\end{table}

\Cref{table:lt} shows that models augmented with VCReg consistently outperform the standard ResNet-32 models on imbalanced datasets. These results are noteworthy because they demonstrate that VCReg effectively enhances the model's ability to discriminate between classes in imbalanced settings. This establishes VCReg as a valuable tool for real-world applications where class imbalance is often a concern.

\subsection{Self-Supervised Learning with VCReg}

Our subsequent investigation focuses on examining the synergy between VCReg and existing self-supervised learning paradigms. 
As mentioned in the previous sections, we apply VCReg not only to the final but also to intermediate representations.
So in all of the following experiments for self-supervised learning with VCReg, we apply the original loss function to the output of the network, and the VCReg loss to all the intermediate representations.

We employ a ResNet-50 architecture, training it for 100 epochs under four different configurations: using either SimCLR loss or VICReg loss, coupled with the ImageNet dataset. For evaluation, we conduct linear probing tests on multiple downstream task datasets, following the protocols prescribed by \cite{misra2020self, zbontar2021barlow}.

\begin{table*}[ht]
\caption{\textbf{Impact of VCReg on Self-Supervised Learning Methods}: This table presents a comparative analysis of ResNet-50 models pretrained with SimCLR and VICReg losses on ImageNet, both with and without the VCReg applied. The models are evaluated using linear probing on various downstream task datasets. The VCReg models consistently outperform the non-VCReg models, showcasing the method's broad utility in transfer learning for self-supervised learning scenarios. Averages are calculated excluding ImageNet results.}
\label{ssl}
\vskip 0.1in
\centering
\scalebox{0.9}{
\begin{tabular}{lcccccccccc}
\toprule
{Pretraining Methods} & {ImageNet} & {iNat18} & {Places} & {Food} & {Cars} & {Aircraft} & {Pets} & {Flowers} & {DTD} & {Average} \\
\midrule
{SimCLR} & \textbf{67.2\%} & 37.2\% & 52.1\% & 66.4\% & 35.7\% & \textbf{62.3\%} & 76.3\% & 82.6\% & 68.1\% & 60.09\% \\
\textbf{SimCLR (VCReg)} & \textbf{67.1\%} & \textbf{41.3\%} & \textbf{52.3\%} & \textbf{67.7\%} & \textbf{40.6\%} & {61.9\%} & \textbf{76.6\%} & \textbf{83.6\%} & \textbf{69.0\%} & \textbf{61.63\%} \\
\midrule
{VICReg} & 65.2\% & \textbf{41.7\%} & 48.2\% & 61.0\% & 27.3\% & 51.2\% & 79.1\% & 74.3\% & 65.4\% & 56.03\% \\
\textbf{VICReg (VCReg)} & \textbf{66.3\%} & {41.4\%} & \textbf{49.6\%} & \textbf{61.6\%} & \textbf{29.3\%} & \textbf{54.2\%} & \textbf{79.7\%} & \textbf{74.5\%} & \textbf{66.5\%} & \textbf{57.10\%} \\
\bottomrule
\end{tabular}
}
\end{table*}

As indicated in Table \ref{ssl}, integrating VCReg into self-supervised learning paradigms such as SimCLR and VICReg results in consistent performance improvements for transfer learning. Specifically, the linear probing accuracies are enhanced across nearly all the evaluated datasets. These gains underscore the broad applicability and versatility of VCReg, demonstrating its potential to enhance various machine learning methodologies.

\subsection{Hierarchical Classification} 
To evaluate the efficacy of the learned representations across multiple levels of class granularity, we conduct experiments on the CIFAR100 dataset as well as five distinct subsets of ImageNet \citep{robustness}. In each dataset, every data sample is tagged with both superclass and subclass labels, denoted as \( (x_i, y^{\mathrm{sup}}_i, y^{\mathrm{sub}}_i) \). Note that while samples sharing the same subclass label also share the same superclass label, the reverse does not necessarily hold true.
Initially, the model is trained using only the superclass labels, i.e., the \( (x_i, y^{\mathrm{sup}}_i) \) pairs. Subsequently, linear probing is employed with the subclass labels \( (x_i, y^{\mathrm{sub}}_i) \) to assess the quality of abstract features at the superclass level.

\begin{table*}[h]
  \caption{\textbf{Impact of VCReg on Hierarchical Classification in ConvNeXt Models}: This table summarizes the classification accuracies obtained with ConvNeXt models, both with and without the VCReg regularization, across multiple datasets featuring hierarchical class structures. The models were initially trained using superclass labels and subsequently probed using subclass labels. VCReg consistently boosts performance in subclass classification tasks.}
  \label{super_sub}
  \vskip 0.1in
  \centering  
  \begin{small}
  \begin{tabular}{lcccccc}
    \toprule
    & & \multicolumn{5}{c}{Subsets of ImageNet} \\
    \cmidrule(r){3-7}
    & CIFAR100 & living\_9 & mixed\_10 & mixed\_13 & geirhos\_16 & big\_12 \\
    \midrule
    Superclass Count & 20 & 9 & 10 & 13 & 16 & 12 \\
    Subclass Count & 100 & 72 & 60 & 78 & 32 & 240 \\
    \midrule
    ConvNeXt & 60.7\% & 53.4\% & 60.3\% & 61.1\% & 60.5\% & 51.8\% \\
    ConvNeXt (VCReg) & \textbf{72.9\%} & \textbf{62.2\%} & \textbf{67.7\%} & \textbf{66.0\%} & \textbf{70.1\%} & \textbf{61.5\%} \\
    \bottomrule
  \end{tabular}
  \end{small}
\end{table*}

Table \ref{super_sub} presents key performance metrics, highlighting the substantial improvements VCReg brings to subclass classification. The improvements are consistent across all datasets, with the CIFAR100 dataset showing the most significant gain—an increase in accuracy from 60.7\% to 72.9\%.
These results underscore VCReg's capability to assist neural networks in generating feature representations that are not only discriminative at the superclass level but are also well-suited for subclass distinctions. This attribute is particularly advantageous in real-world applications where class categorizations often exist within a hierarchical framework.

\section{Exploring the Benefits of VCReg}

This section aims to thoroughly unpack the multi-faceted benefits of VCReg in the context of supervised neural network training. Specifically, we discuss its capability to address challenges such as gradient starvation \citep{pezeshki2021gradient}, neural collapse \citep{papyan2020prevalence}, noisy data, and the preservation of information richness during model training \citep{shwartz2022information}.

\subsection{Mitigating Gradient Starvation}

\begin{figure}[t]
\centering
\includegraphics[height=34mm, width=60mm]{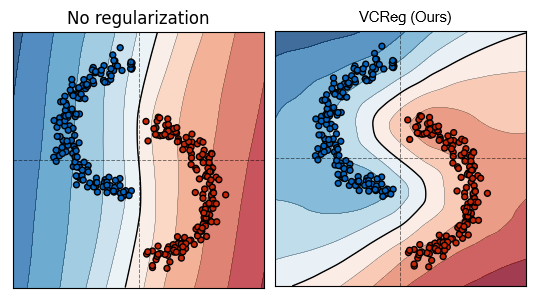}
\caption{\textbf{Comparative evaluation between training with and without VCReg on a ``Two-Moon'' Synthetic Dataset.} Decision boundaries are averaged over ten distinct runs with random data point sampling and model initialization. A single run's data points are displayed for visual clarity. The contrast between VCReg and ``No regularization'' underscores the latter's limitations in forming intricate decision boundaries, while highlighting VCReg's effectiveness in generating meaningful ones.}

\label{fig:two_moon}
\end{figure}

In line with the original study on gradient starvation \citep{pezeshki2021gradient}, we observe that most traditional regularization techniques fall short of capturing the vital features for the ``two-moon'' dataset experiment. To assess the effectiveness of VCReg, we replicate this setting with a three-layer network and apply our method during training. Our visualized results in \Cref{fig:two_moon} make it apparent that VCReg has a marked advantage over traditional regularization techniques, particularly in the aspects of separation margins.  Thus, it is reasonable to conclude that VCReg can help mitigate gradient starvation. Please check section \ref{tmsection} for the detailed information about experiments related to the ``two-moon'' dataset.

\subsection{Preventing Neural Collapse and Information Compression}

To deepen our understanding of VCReg and its training dynamics, we closely examine its learned representations. A recent study \citep{papyan2020prevalence} observed a peculiar trend in deep networks trained for classification tasks: the top-layer feature embeddings of training samples from the same class tend to cluster around their respective class means, which are as distant from each other as possible. However, this phenomenon could potentially result in a loss of diversity among the learned features \citep{papyan2020prevalence}, thus curtailing the network's capacity to grasp the complexity of the data and leading to suboptimal performance for transfer learning  \citep{li2018measuring}.

Our neural collapse investigation includes two key metrics:

\textbf{Class-Distance Normalized Variance (CDNV)} For a feature map \( f:\mathbb{R}^d \to \mathbb{R}^p \) and two unlabeled sets of samples \( S_1,S_2 \subset \mathbb{R}^{d} \), the CDNV is defined as
% \begin{align}
% V_f(S_1,S_2) = \frac{\Var_f(S_1) + \Var_f(S_2)}{2\|\mu_f(S_1)-\mu_f(S_2)\|^2}, 
% \end{align}
\begin{align}
V_f(S_1,S_2) = \frac{\sigma^2_f(S_1) + \sigma^2_f(S_2)}{2\|\mu_f(S_1)-\mu_f(S_2)\|^2}, 
\end{align}
% where \( \mu_f(S) \) and \( \Var_f(S) \) signify the mean and variance of the set \( \{f(x) \mid x \in S\} \). 
where \( \mu_f(S) \) and \( \sigma^2_f(S) \) signify the mean and variance of the set \( \{f(x) \mid x \in S\} \). 
This metric measures the degree of clustering of the features extracted from \( S_1 \) and \( S_2 \), in relation to the distance between their respective features. A value approaching zero indicates perfect clustering.

\textbf{Nearest Class-Center Classifier (NCC)} This classifier is defined as 
\begin{align}
\hat{h}(x) = \argmin_{c\in [C]} \|f(x) - \mu_{f}(S_c)\|
\end{align}
According to this measure, during training, collapsed feature embeddings in the penultimate layer become separable, and the classifier converges to the ``nearest class-center classifier''.

\textbf{Preventing Information Compression} Although effective compression often yields superior representations, overly aggressive compression might cause the loss of crucial information about the target task \citep{shwartz2018representation,shwartz2020information, 2023arXiv230409355S}.
To investigate the compression during the learning, we use the mutual information neural estimation (MINE)~\citep{belghazi2018mine}, a method specifically designed to estimate the mutual information between the input and its corresponding embedded representation. This metric effectively gauges the complexity level of the representation, essentially indicating how much information (in terms of number of bits) it encodes.

We evaluate the learned representations of two ConvNeXt models \citep{liu2022convnet},  which are trained on ImageNet with supervised loss. One model is trained with VCReg, while the other is trained without VCReg.
As demonstrated in Table \ref{ta:collapse}, both types of collapse, measured by CDNV and NCC, and the mutual information estimation reveal that VCReg representations have significantly more diverse features (lower neural collapse) and contain more information compared to regular training. This suggests that not only does VCReg achieve superior results, but it also yields representations which contain more information.

In summary, the VCReg method mitigates the neural collapse phenomenon and prevents excessive information compression, two crucial factors that often limit the effectiveness of deep learning models in transfer learning tasks. Our findings highlight the potential of VCReg as a valuable addition to the deep learning toolbox, significantly increasing the generalizability of learned representations.

\begin{table}[t]
  \caption{\textbf{VCReg learns richer representation and prevents neural collapse and information compression} Metrics include Class-Distance Normalized Variance (CDNV), Nearest Class-Center Classifier (NCC), and Mutual Information (MI). Higher values in each metric for the VCReg model indicate reduced neural collapse and richer feature representations.}
  \vskip 0.1in
  \label{ta:collapse}
  \centering
  \scalebox{0.90}{
  \begin{tabular}{lccc}
    \toprule
    Network & CDNV & NCC & MI \\
    \midrule
    ConvNeXt & 0.28 &  0.99 & 2.8  \\
    ConvNeXt (VCReg) & \textbf{0.56} &   $ \textbf{0.81}$ & $\textbf{4.6}$ \\
    \bottomrule
  \end{tabular}}
\end{table}
\subsection{Providing Robustness to Noise}

In real-world scenarios, encountering noise is a common challenge, making robustness against noise a crucial feature for any effective transfer learning algorithm. Recognizing the ubiquity of noise in practical applications, we aim to evaluate the capability of VCReg to bolster transfer learning performance in noisy environments.

\begin{figure}[ht]
    \centering
    \includegraphics[width=0.5\textwidth]{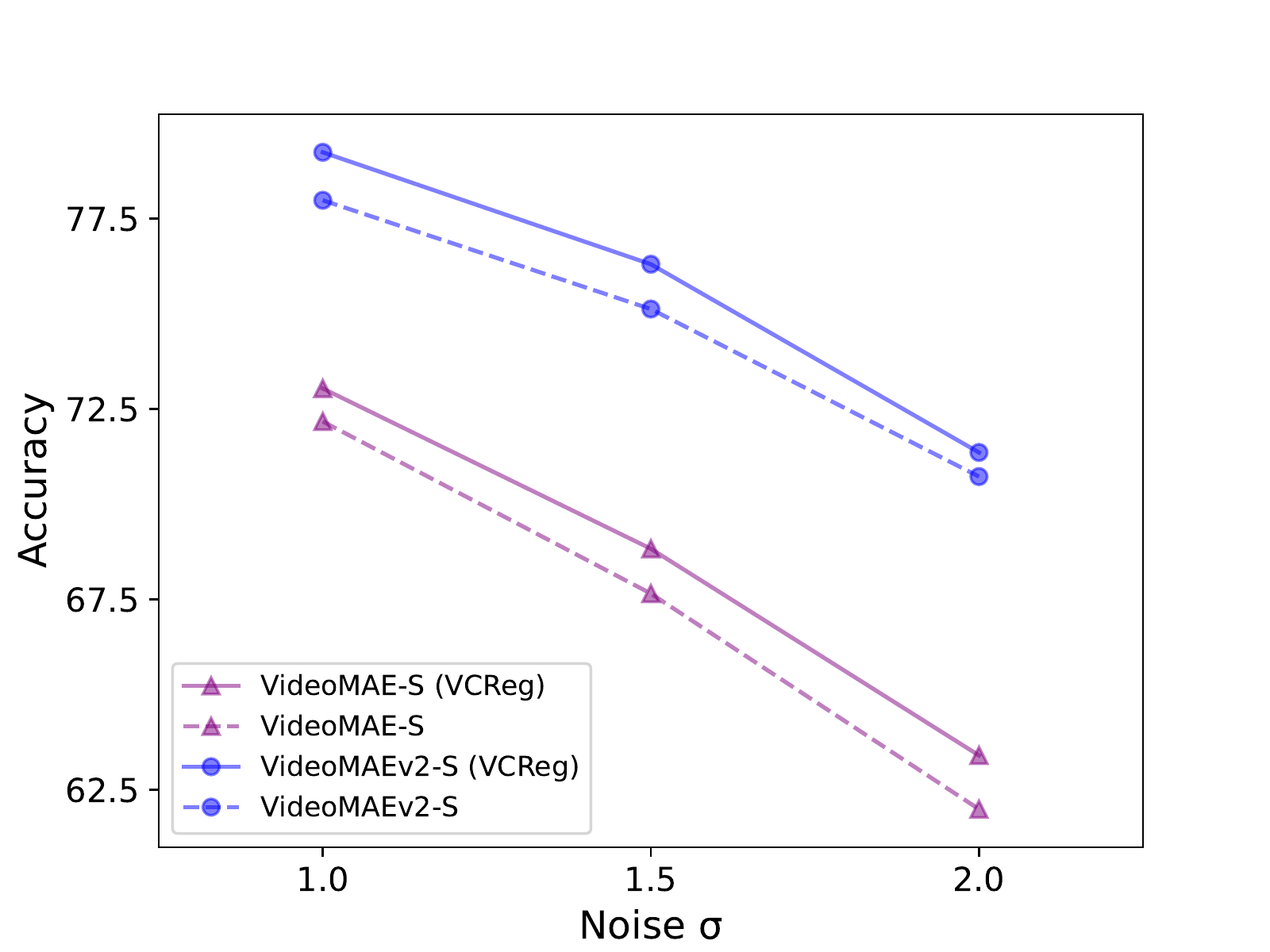}
    \caption{\textbf{Impact of VCReg amidst noisy data}: This figure shows the top-1 accuracy of VideoMAE-S and VideoMAEv2-S when fine-tuned for action recognition using HMDB51 corrupted with synthetic noise. We corrupt the data with Gaussian noise with standard deviation $\sigma\in\{1, 1.5, 2\}$. Models with VCReg outperform their non-regularized counterparts in this setting. }
    \label{fig:transfer-learning-noise}
\end{figure}

For this purpose, we utilize video networks initially pretrained on Kinetics-400 and Kinetics-710, as mentioned in section \ref{sec:transfer-with-video}. We then finetune these networks on the HMDB51 dataset, which is deliberately subjected to varying levels of Gaussian noise. The findings in Table \ref{fig:transfer-learning-noise} reveal a clear advantage: incorporating VCReg notably improves the resilience of VideoMAE-S and VideoMAEv2-S models to noisy data, a robustness not observed in models without VCReg. This trend of increased durability against noise is consistently seen in larger models, such as VideoMAE-B and VideoMAEv2-B. For a more granular analysis, Appendix \ref{app:robustness-noise}  provides a thorough description of the results, complete with detailed figures.

This investigation highlights the necessity of achieving optimal performance in non-ideal settings. It emphasizes the critical need for maintaining robustness and reliability under the challenges commonly encountered in real-world settings, such as noise. This dual capability significantly boosts a model's practical value and reliability.

% Finally, we investigate whether VCReg can enhance transfer learning performance in the presence of noise. For this purpose, we use networks pre-trained on Kinetics-400 and Kinetics-710 from section \ref{sec:transfer-with-video} and fine-tune them on HMDB51 corrupted with different levels of Gaussian noise.  Figure \ref{fig:transfer-learning-noise} shows that VideoMAE-S (VCReg) and VideoMAEv2-S (VCReg) models are more robust to noisy data than their non-regularized counterparts. A  similar trend holds for VideoMAE-B (VCReg) and VideoMAEv2-B (VCReg) (please refer to Figures \ref{fig:transfer-learning-noise-VideoMAEB} and \ref{fig:transfer-learning-noise-VideoMAEv2B} in Appendix \ref{app:robustness-noise}.)
 
\section{Conclusion}

In this work, we addressed prevalent challenges in supervised pretraining for transfer learning by introducing Variance-Covariance Regularization (VCReg). Building on the regularization technique of the self-supervised VICReg method, VCReg is designed to cultivate robust and generalizable features. Unlike conventional methods that attach regularization only to the final layer, we efficiently incorporate VCReg across intermediate layers to optimize its efficacy.

Our key contributions are threefold: 
\begin{enumerate}
    \item We present a computationally efficient VCReg implementation that can be adapted to various network architectures.
    \item We provide empirical evidence through comprehensive evaluations on multiple benchmarks, demonstrating that using VCReg yields significant improvements in transfer learning performance across various network architectures and different learning paradigms, including video and image classification, long tail learning, and hierarchical classification.
    \item Our in-depth analyses confirm VCReg's effectiveness in overcoming typical transfer learning hurdles such as neural collapse, gradient starvation, and noise.
\end{enumerate} 

To conclude, VCReg stands out as a potent and adaptable regularization technique that elevates the quality and applicability of learned representations. It enhances both the performance and reliability of models in transfer learning settings and paves the way for further research to achieve highly optimized and generalizable machine learning models.

\section*{Acknowledgements}
This material is partially based upon work supported by the National Science Foundation under NSF Award 1922658.

\bibliography{example_paper}

\begin{thebibliography}{56}
\providecommand{\natexlab}[1]{#1}
\providecommand{\url}[1]{\texttt{#1}}
\expandafter\ifx\csname urlstyle\endcsname\relax
  \providecommand{\doi}[1]{doi: #1}\else
  \providecommand{\doi}{doi: \begingroup \urlstyle{rm}\Url}\fi

\bibitem[Arnab et~al.(2021)Arnab, Dehghani, Heigold, Sun, Lu{\v{c}}i{\'c}, and Schmid]{arnab2021vivit}
Arnab, A., Dehghani, M., Heigold, G., Sun, C., Lu{\v{c}}i{\'c}, M., and Schmid, C.
\newblock Vivit: A video vision transformer.
\newblock In \emph{Proceedings of the IEEE/CVF international conference on computer vision}, pp.\  6836--6846, 2021.

\bibitem[Ayinde et~al.(2019)Ayinde, Inanc, and Zurada]{ayinde2019regularizing}
Ayinde, B.~O., Inanc, T., and Zurada, J.~M.
\newblock Regularizing deep neural networks by enhancing diversity in feature extraction.
\newblock \emph{IEEE transactions on neural networks and learning systems}, 30\penalty0 (9):\penalty0 2650--2661, 2019.

\bibitem[Bardes et~al.(2021)Bardes, Ponce, and LeCun]{bardes2021vicreg}
Bardes, A., Ponce, J., and LeCun, Y.
\newblock Vicreg: Variance-invariance-covariance regularization for self-supervised learning.
\newblock \emph{arXiv preprint arXiv:2105.04906}, 2021.

\bibitem[Belghazi et~al.(2018)Belghazi, Baratin, Rajeswar, Ozair, Bengio, Courville, and Hjelm]{belghazi2018mine}
Belghazi, M.~I., Baratin, A., Rajeswar, S., Ozair, S., Bengio, Y., Courville, A., and Hjelm, R.~D.
\newblock Mine: mutual information neural estimation.
\newblock \emph{arXiv preprint arXiv:1801.04062}, 2018.

\bibitem[Ben-Shaul et~al.(2023)Ben-Shaul, Shwartz-Ziv, Galanti, Dekel, and LeCun]{ben2023reverse}
Ben-Shaul, I., Shwartz-Ziv, R., Galanti, T., Dekel, S., and LeCun, Y.
\newblock Reverse engineering self-supervised learning.
\newblock \emph{arXiv preprint arXiv:2305.15614}, 2023.

\bibitem[Bengio(2012)]{bengio2012deep}
Bengio, Y.
\newblock Deep learning of representations for unsupervised and transfer learning.
\newblock In \emph{Proceedings of ICML workshop on unsupervised and transfer learning}, pp.\  17--36. JMLR Workshop and Conference Proceedings, 2012.

\bibitem[Bommasani et~al.(2021)Bommasani, Hudson, Adeli, Altman, Arora, von Arx, Bernstein, Bohg, Bosselut, Brunskill, et~al.]{bommasani2021opportunities}
Bommasani, R., Hudson, D.~A., Adeli, E., Altman, R., Arora, S., von Arx, S., Bernstein, M.~S., Bohg, J., Bosselut, A., Brunskill, E., et~al.
\newblock On the opportunities and risks of foundation models.
\newblock \emph{arXiv preprint arXiv:2108.07258}, 2021.

\bibitem[Bossard et~al.(2014)Bossard, Guillaumin, and Van~Gool]{bossard2014food}
Bossard, L., Guillaumin, M., and Van~Gool, L.
\newblock Food-101--mining discriminative components with random forests.
\newblock In \emph{Computer Vision--ECCV 2014: 13th European Conference, Zurich, Switzerland, September 6-12, 2014, Proceedings, Part VI 13}, pp.\  446--461. Springer, 2014.

\bibitem[Caruana(1997)]{caruana1997multitask}
Caruana, R.
\newblock Multitask learning.
\newblock \emph{Machine learning}, 28:\penalty0 41--75, 1997.

\bibitem[Chen et~al.(2020)Chen, Kornblith, Norouzi, and Hinton]{chen2020simple}
Chen, T., Kornblith, S., Norouzi, M., and Hinton, G.
\newblock A simple framework for contrastive learning of visual representations.
\newblock In \emph{International conference on machine learning}, pp.\  1597--1607. PMLR, 2020.

\bibitem[Cimpoi et~al.(2014)Cimpoi, Maji, Kokkinos, Mohamed, and Vedaldi]{cimpoi2014describing}
Cimpoi, M., Maji, S., Kokkinos, I., Mohamed, S., and Vedaldi, A.
\newblock Describing textures in the wild.
\newblock In \emph{Proceedings of the IEEE conference on computer vision and pattern recognition}, pp.\  3606--3613, 2014.

\bibitem[Cogswell et~al.(2015)Cogswell, Ahmed, Girshick, Zitnick, and Batra]{cogswell2015reducing}
Cogswell, M., Ahmed, F., Girshick, R., Zitnick, L., and Batra, D.
\newblock Reducing overfitting in deep networks by decorrelating representations.
\newblock \emph{arXiv preprint arXiv:1511.06068}, 2015.

\bibitem[Deng et~al.(2009)Deng, Dong, Socher, Li, Li, and Fei-Fei]{deng2009imagenet}
Deng, J., Dong, W., Socher, R., Li, L.-J., Li, K., and Fei-Fei, L.
\newblock Imagenet: A large-scale hierarchical image database.
\newblock In \emph{2009 IEEE conference on computer vision and pattern recognition}, pp.\  248--255. Ieee, 2009.

\bibitem[Dosovitskiy et~al.(2020)Dosovitskiy, Beyer, Kolesnikov, Weissenborn, Zhai, Unterthiner, Dehghani, Minderer, Heigold, Gelly, et~al.]{dosovitskiy2020image}
Dosovitskiy, A., Beyer, L., Kolesnikov, A., Weissenborn, D., Zhai, X., Unterthiner, T., Dehghani, M., Minderer, M., Heigold, G., Gelly, S., et~al.
\newblock An image is worth 16x16 words: Transformers for image recognition at scale.
\newblock \emph{arXiv preprint arXiv:2010.11929}, 2020.

\bibitem[Engstrom et~al.(2019)Engstrom, Ilyas, Santurkar, and Tsipras]{robustness}
Engstrom, L., Ilyas, A., Santurkar, S., and Tsipras, D.
\newblock Robustness (python library), 2019.
\newblock URL \url{https://github.com/MadryLab/robustness}.

\bibitem[Geiping et~al.(2022)Geiping, Goldblum, Somepalli, Shwartz-Ziv, Goldstein, and Wilson]{geiping2022much}
Geiping, J., Goldblum, M., Somepalli, G., Shwartz-Ziv, R., Goldstein, T., and Wilson, A.~G.
\newblock How much data are augmentations worth? an investigation into scaling laws, invariance, and implicit regularization.
\newblock \emph{arXiv preprint arXiv:2210.06441}, 2022.

\bibitem[He et~al.(2016)He, Zhang, Ren, and Sun]{he2016deep}
He, K., Zhang, X., Ren, S., and Sun, J.
\newblock Deep residual learning for image recognition.
\newblock In \emph{Proceedings of the IEEE conference on computer vision and pattern recognition}, pp.\  770--778, 2016.

\bibitem[Hinton et~al.(2012)Hinton, Srivastava, Krizhevsky, Sutskever, and Salakhutdinov]{hinton2012improving}
Hinton, G.~E., Srivastava, N., Krizhevsky, A., Sutskever, I., and Salakhutdinov, R.~R.
\newblock Improving neural networks by preventing co-adaptation of feature detectors.
\newblock \emph{arXiv preprint arXiv:1207.0580}, 2012.

\bibitem[Ioffe \& Szegedy(2015)Ioffe and Szegedy]{ioffe2015batch}
Ioffe, S. and Szegedy, C.
\newblock Batch normalization: Accelerating deep network training by reducing internal covariate shift.
\newblock In \emph{International conference on machine learning}, pp.\  448--456. pmlr, 2015.

\bibitem[Kay et~al.(2017)Kay, Carreira, Simonyan, Zhang, Hillier, Vijayanarasimhan, Viola, Green, Back, Natsev, et~al.]{kay2017kinetics}
Kay, W., Carreira, J., Simonyan, K., Zhang, B., Hillier, C., Vijayanarasimhan, S., Viola, F., Green, T., Back, T., Natsev, P., et~al.
\newblock The kinetics human action video dataset.
\newblock \emph{arXiv preprint arXiv:1705.06950}, 2017.

\bibitem[Kessy et~al.(2018)Kessy, Lewin, and Strimmer]{kessy2018optimal}
Kessy, A., Lewin, A., and Strimmer, K.
\newblock Optimal whitening and decorrelation.
\newblock \emph{The American Statistician}, 72\penalty0 (4):\penalty0 309--314, 2018.

\bibitem[Kornblith et~al.(2021)Kornblith, Chen, Lee, and Norouzi]{kornblith2021better}
Kornblith, S., Chen, T., Lee, H., and Norouzi, M.
\newblock Why do better loss functions lead to less transferable features?
\newblock \emph{Advances in Neural Information Processing Systems}, 34:\penalty0 28648--28662, 2021.

\bibitem[Krause et~al.(2013)Krause, Stark, Deng, and Fei-Fei]{krause20133d}
Krause, J., Stark, M., Deng, J., and Fei-Fei, L.
\newblock 3d object representations for fine-grained categorization.
\newblock In \emph{Proceedings of the IEEE international conference on computer vision workshops}, pp.\  554--561, 2013.

\bibitem[Krizhevsky et~al.(2009)Krizhevsky, Hinton, et~al.]{krizhevsky2009learning}
Krizhevsky, A., Hinton, G., et~al.
\newblock Learning multiple layers of features from tiny images.
\newblock 2009.

\bibitem[Kuehne et~al.(2011)Kuehne, Jhuang, Garrote, Poggio, and Serre]{kuehne2011hmdb}
Kuehne, H., Jhuang, H., Garrote, E., Poggio, T., and Serre, T.
\newblock Hmdb: a large video database for human motion recognition.
\newblock In \emph{2011 International conference on computer vision}, pp.\  2556--2563. IEEE, 2011.

\bibitem[Laakom et~al.(2023)Laakom, Raitoharju, Iosifidis, and Gabbouj]{laakom2023wld}
Laakom, F., Raitoharju, J., Iosifidis, A., and Gabbouj, M.
\newblock Wld-reg: A data-dependent within-layer diversity regularizer.
\newblock \emph{arXiv preprint arXiv:2301.01352}, 2023.

\bibitem[LeCun et~al.(2002)LeCun, Bottou, Orr, and M{\"u}ller]{lecun2002efficient}
LeCun, Y., Bottou, L., Orr, G.~B., and M{\"u}ller, K.-R.
\newblock Efficient backprop.
\newblock In \emph{Neural networks: Tricks of the trade}, pp.\  9--50. Springer, 2002.

\bibitem[Li et~al.(2018)Li, Farkhoor, Liu, and Yosinski]{li2018measuring}
Li, C., Farkhoor, H., Liu, R., and Yosinski, J.
\newblock Measuring the intrinsic dimension of objective landscapes.
\newblock \emph{arXiv preprint arXiv:1804.08838}, 2018.

\bibitem[Li et~al.(2022)Li, Wang, He, Li, Wang, Wang, and Qiao]{li2022uniformerv2}
Li, K., Wang, Y., He, Y., Li, Y., Wang, Y., Wang, L., and Qiao, Y.
\newblock Uniformerv2: Spatiotemporal learning by arming image vits with video uniformer, 2022.

\bibitem[Liu et~al.(2022)Liu, Mao, Wu, Feichtenhofer, Darrell, and Xie]{liu2022convnet}
Liu, Z., Mao, H., Wu, C.-Y., Feichtenhofer, C., Darrell, T., and Xie, S.
\newblock A convnet for the 2020s.
\newblock In \emph{Proceedings of the IEEE/CVF Conference on Computer Vision and Pattern Recognition}, pp.\  11976--11986, 2022.

\bibitem[Maji et~al.(2013)Maji, Kannala, Rahtu, Blaschko, and Vedaldi]{maji13fine-grained}
Maji, S., Kannala, J., Rahtu, E., Blaschko, M., and Vedaldi, A.
\newblock Fine-grained visual classification of aircraft.
\newblock Technical report, 2013.

\bibitem[Misra \& Maaten(2020)Misra and Maaten]{misra2020self}
Misra, I. and Maaten, L. v.~d.
\newblock Self-supervised learning of pretext-invariant representations.
\newblock In \emph{Proceedings of the IEEE/CVF conference on computer vision and pattern recognition}, pp.\  6707--6717, 2020.

\bibitem[Neyshabur et~al.(2017)Neyshabur, Bhojanapalli, McAllester, and Srebro]{neyshabur2017exploring}
Neyshabur, B., Bhojanapalli, S., McAllester, D., and Srebro, N.
\newblock Exploring generalization in deep learning.
\newblock \emph{Advances in neural information processing systems}, 30, 2017.

\bibitem[Nilsback \& Zisserman(2008)Nilsback and Zisserman]{nilsback2008automated}
Nilsback, M.-E. and Zisserman, A.
\newblock Automated flower classification over a large number of classes.
\newblock In \emph{2008 Sixth Indian Conference on Computer Vision, Graphics \& Image Processing}, pp.\  722--729. IEEE, 2008.

\bibitem[Pan \& Yang(2010)Pan and Yang]{pan2010survey}
Pan, S.~J. and Yang, Q.
\newblock A survey on transfer learning.
\newblock \emph{IEEE Transactions on knowledge and data engineering}, 22\penalty0 (10):\penalty0 1345--1359, 2010.

\bibitem[Papyan et~al.(2020)Papyan, Han, and Donoho]{papyan2020prevalence}
Papyan, V., Han, X., and Donoho, D.~L.
\newblock Prevalence of neural collapse during the terminal phase of deep learning training.
\newblock \emph{Proceedings of the National Academy of Sciences}, 117\penalty0 (40):\penalty0 24652--24663, 2020.

\bibitem[Parkhi et~al.(2012)Parkhi, Vedaldi, Zisserman, and Jawahar]{parkhi2012cats}
Parkhi, O.~M., Vedaldi, A., Zisserman, A., and Jawahar, C.
\newblock Cats and dogs.
\newblock In \emph{2012 IEEE conference on computer vision and pattern recognition}, pp.\  3498--3505. IEEE, 2012.

\bibitem[Paszke et~al.(2019)Paszke, Gross, Massa, Lerer, Bradbury, Chanan, Killeen, Lin, Gimelshein, Antiga, Desmaison, Kopf, Yang, DeVito, Raison, Tejani, Chilamkurthy, Steiner, Fang, Bai, and Chintala]{NEURIPS2019_9015}
Paszke, A., Gross, S., Massa, F., Lerer, A., Bradbury, J., Chanan, G., Killeen, T., Lin, Z., Gimelshein, N., Antiga, L., Desmaison, A., Kopf, A., Yang, E., DeVito, Z., Raison, M., Tejani, A., Chilamkurthy, S., Steiner, B., Fang, L., Bai, J., and Chintala, S.
\newblock Pytorch: An imperative style, high-performance deep learning library.
\newblock In \emph{Advances in Neural Information Processing Systems 32}, pp.\  8024--8035. Curran Associates, Inc., 2019.
\newblock URL \url{http://papers.neurips.cc/paper/9015-pytorch-an-imperative-style-high-performance-deep-learning-library.pdf}.

\bibitem[Pezeshki et~al.(2021)Pezeshki, Kaba, Bengio, Courville, Precup, and Lajoie]{pezeshki2021gradient}
Pezeshki, M., Kaba, O., Bengio, Y., Courville, A.~C., Precup, D., and Lajoie, G.
\newblock Gradient starvation: A learning proclivity in neural networks.
\newblock \emph{Advances in Neural Information Processing Systems}, 34:\penalty0 1256--1272, 2021.

\bibitem[Shwartz-Ziv(2022)]{shwartz2022information}
Shwartz-Ziv, R.
\newblock Information flow in deep neural networks.
\newblock \emph{arXiv preprint arXiv:2202.06749}, 2022.

\bibitem[Shwartz-Ziv \& Alemi(2020)Shwartz-Ziv and Alemi]{shwartz2020information}
Shwartz-Ziv, R. and Alemi, A.~A.
\newblock Information in infinite ensembles of infinitely-wide neural networks.
\newblock In \emph{Symposium on Advances in Approximate Bayesian Inference}, pp.\  1--17. PMLR, 2020.

\bibitem[Shwartz-Ziv \& LeCun(2023)Shwartz-Ziv and LeCun]{2023arXiv230409355S}
Shwartz-Ziv, R. and LeCun, Y.
\newblock To compress or not to compress--self-supervised learning and information theory: A review.
\newblock \emph{arXiv preprint arXiv:2304.09355}, 2023.

\bibitem[Shwartz-Ziv et~al.(2018)Shwartz-Ziv, Painsky, and Tishby]{shwartz2018representation}
Shwartz-Ziv, R., Painsky, A., and Tishby, N.
\newblock Representation compression and generalization in deep neural networks, 2018.

\bibitem[Shwartz-Ziv et~al.(2022)Shwartz-Ziv, Balestriero, and LeCun]{shwartz2022we}
Shwartz-Ziv, R., Balestriero, R., and LeCun, Y.
\newblock What do we maximize in self-supervised learning?
\newblock \emph{arXiv preprint arXiv:2207.10081}, 2022.

\bibitem[Shwartz-Ziv et~al.(2023)Shwartz-Ziv, Balestriero, Kawaguchi, Rudner, and LeCun]{shwartz2023information}
Shwartz-Ziv, R., Balestriero, R., Kawaguchi, K., Rudner, T.~G., and LeCun, Y.
\newblock An information-theoretic perspective on variance-invariance-covariance regularization.
\newblock \emph{arXiv preprint arXiv:2303.00633}, 2023.

\bibitem[Simonyan \& Zisserman(2014)Simonyan and Zisserman]{simonyan2014two}
Simonyan, K. and Zisserman, A.
\newblock Two-stream convolutional networks for action recognition in videos.
\newblock \emph{Advances in neural information processing systems}, 27, 2014.

\bibitem[Tong et~al.(2022)Tong, Song, Wang, and Wang]{tong2022videomae}
Tong, Z., Song, Y., Wang, J., and Wang, L.
\newblock Videomae: Masked autoencoders are data-efficient learners for self-supervised video pre-training.
\newblock \emph{Advances in neural information processing systems}, 35:\penalty0 10078--10093, 2022.

\bibitem[Van~Horn et~al.(2018)Van~Horn, Mac~Aodha, Song, Cui, Sun, Shepard, Adam, Perona, and Belongie]{van2018inaturalist}
Van~Horn, G., Mac~Aodha, O., Song, Y., Cui, Y., Sun, C., Shepard, A., Adam, H., Perona, P., and Belongie, S.
\newblock The inaturalist species classification and detection dataset.
\newblock In \emph{Proceedings of the IEEE conference on computer vision and pattern recognition}, pp.\  8769--8778, 2018.

\bibitem[Wang et~al.(2023)Wang, Huang, Zhao, Tong, He, Wang, Wang, and Qiao]{wang2023videomae}
Wang, L., Huang, B., Zhao, Z., Tong, Z., He, Y., Wang, Y., Wang, Y., and Qiao, Y.
\newblock Videomae v2: Scaling video masked autoencoders with dual masking.
\newblock In \emph{Proceedings of the IEEE/CVF Conference on Computer Vision and Pattern Recognition}, pp.\  14549--14560, 2023.

\bibitem[Weiss et~al.(2016)Weiss, Khoshgoftaar, and Wang]{Weiss2016ASO}
Weiss, K.~R., Khoshgoftaar, T.~M., and Wang, D.
\newblock A survey of transfer learning.
\newblock \emph{Journal of Big Data}, 3, 2016.

\bibitem[Yosinski et~al.(2014)Yosinski, Clune, Bengio, and Lipson]{yosinski2014transferable}
Yosinski, J., Clune, J., Bengio, Y., and Lipson, H.
\newblock How transferable are features in deep neural networks?
\newblock \emph{Advances in neural information processing systems}, 27, 2014.

\bibitem[Zbontar et~al.(2021)Zbontar, Jing, Misra, LeCun, and Deny]{zbontar2021barlow}
Zbontar, J., Jing, L., Misra, I., LeCun, Y., and Deny, S.
\newblock Barlow twins: Self-supervised learning via redundancy reduction.
\newblock \emph{arXiv preprint arXiv:2103.03230}, 2021.

\bibitem[Zhang et~al.(2016)Zhang, Bengio, Hardt, Recht, and Vinyals]{zhang2016understanding}
Zhang, C., Bengio, S., Hardt, M., Recht, B., and Vinyals, O.
\newblock Understanding deep learning requires rethinking generalization. corr abs/1611.03530 (2016).
\newblock \emph{arXiv preprint arxiv:1611.03530}, 2016.

\bibitem[Zhang et~al.(2021)Zhang, Bengio, Hardt, Recht, and Vinyals]{zhang2021understanding}
Zhang, C., Bengio, S., Hardt, M., Recht, B., and Vinyals, O.
\newblock Understanding deep learning (still) requires rethinking generalization.
\newblock \emph{Communications of the ACM}, 64\penalty0 (3):\penalty0 107--115, 2021.

\bibitem[Zhou et~al.(2014)Zhou, Lapedriza, Xiao, Torralba, and Oliva]{zhou2014learning}
Zhou, B., Lapedriza, A., Xiao, J., Torralba, A., and Oliva, A.
\newblock Learning deep features for scene recognition using places database.
\newblock \emph{Advances in neural information processing systems}, 27, 2014.

\bibitem[Zhuang et~al.(2020)Zhuang, Qi, Duan, Xi, Zhu, Zhu, Xiong, and He]{zhuang2020comprehensive}
Zhuang, F., Qi, Z., Duan, K., Xi, D., Zhu, Y., Zhu, H., Xiong, H., and He, Q.
\newblock A comprehensive survey on transfer learning.
\newblock \emph{Proceedings of the IEEE}, 109\penalty0 (1):\penalty0 43--76, 2020.

\end{thebibliography}
\bibliographystyle{icml2024}

\appendix
\newpage
\appendix

\section{Experimental Investigation on Effective Application of VCReg to Standard Networks}
\label{investigation}

To determine the optimal manner of integrating the VCReg into a standard network, we conducted several experiments utilizing the ConvNeXt-Atto architecture, trained on ImageNet following the torchvision \citep{NEURIPS2019_9015} training recipe. To reduce the training time, we limited the network training to 90 epochs with a batch size of 4096.
The complete configuration comprised 90 epochs, a batch size of 4096, two learning rate of $\{0.016, 0.008\}$ with a 5 epochs linear warmup followed by a cosine annealing decay. The weight decay was set at $0.05$ and the norm layers were excluded from the weight decay. we experimented with $\alpha \in \{1.28, 0.64, 0.32, 0.16\}$ and $\beta \in \{0.16, 0.08, 0.04, 0.02, 0.01\}$.

We experimented with incorporating the VCReg layers in four different locations:
\begin{enumerate}
\item Applying the VCReg exclusively to the second last representation (the input of the classification layer).
\item Applying VCReg to the output of each ConvNeXt block.
\item Applying VCReg to the output of each downsample layer.
\item Applying VCReg to the output of both, each ConvNeXt block and each downsample layer.
\end{enumerate}

The VCReg layer was implemented as detailed in \ref{alg:vcr_algorithm}, with the addition of a mean removal layer along the batch preceding the VCReg layer to ensure that the VCReg input exhibited a zero mean.

\begin{table*}[h]
\caption{\textbf{Transfer Learning Experiments with Different VCReg Configurations}}
\label{transfer_learning_5}
\centering
\begin{tabular}{lllllll}
\toprule
Architecture & Food & Cars & Aircraft & Pets & Flowers & DTD \\
\midrule
ConvNeXt-Atto (VCReg1) & 63.2\% & 39.6\% & 55.9\% & 89.1\% & 85.3\% & 65.1\% \\
ConvNeXt-Atto (VCReg2) & \textbf{66.8}\% & 48.1\% & \textbf{60.4}\% & \textbf{91.1}\% & \textbf{86.4}\% & \textbf{66.4}\% \\
ConvNeXt-Atto (VCReg3) & 64.0\% & 40.9\% & 56.5\% & 89.4\% & 85.9\% & 65.1\% \\
ConvNeXt-Atto (VCReg4) & 66.7\% & \textbf{48.3}\% & 59.6\% & 90.6\% & 85.6\% & 66.1\% \\
\bottomrule
\end{tabular}
\end{table*}

The results in Table \ref{transfer_learning_5} indicate superior performance when the VCReg layer is applied to the output of each block (second setup) 
 or applied to the output of blocks and downsample layers (fourth setup) compared to the other setups. Considering architectures like ViT lack downsample layers, for consistency across different architectures, we decided to use the second configuration for further experiments.

\section{The Fast Implementation of the VCReg}
\label{sec:appendix-fast}
The VCRegeg does not affect the forward pass in any way, allowing us to substantially speed up the implementation by modifying the backward function directly. Instead of computing the VCReg loss and backpropagating it, we can directly alter the calculated gradient. This is possible since the VCReg loss calculation only requires the current representation. The specifics of this speed-optimized implementation are outlined in Algorithm \ref{alg:vcr_algorithm}.

\begin{algorithm*}[t]
   \caption{PyTorch-Style Pseudocode for Fast VCReg Implementation}
   \label{alg:vcr_algorithm}
   
    \definecolor{codeblue}{rgb}{0.25,0.5,0.5}
    \lstset{
      basicstyle=\fontsize{7.2pt}{7.2pt}\ttfamily\bfseries,
      commentstyle=\fontsize{7.2pt}{7.2pt}\color{codeblue},
      keywordstyle=\fontsize{7.2pt}{7.2pt},
    }
\begin{lstlisting}[language=python, texcl=true, mathescape=true]
# $\alpha$, $\beta$ and $\epsilon$ : hyperparameters
# mm: matrix-matrix multiplication

class VarianceCovarianceRegularizationFunction(Function):
    # forward pass
    # We assume the input has zero mean per channel
    # In practice, we apply a batch demean operation before calling the function 
    def forward(ctx, input):
        ctx.save_for_backward(input)
        return input
    # backward pass
    def backward(ctx, grad_output):
        input, = ctx.saved_tensors
        # reshape the input to have (n, d) shape
        flattened_input = input.flatten(start_dim=0, end_dim=-2)
        n, d = flattened_input.shape
        # calculate the covariance matrix
        covariance_matrix = mm(flattened_input.t(), flattened_input) / (n - 1)
        # calculate the gradient
        diagonal = F.threshold(rsqrt(covariance_matrix.diagonal() + \epsilon), 1.0, 0.0)
        std_grad_input = diagonal * flattened_input
        cov_grad_input = torch.mm(flattened_input, covariance_matrix.fill_diagonal_(0))
        
        grad_input = grad_output
                    - $\alpha / (d(n-1))$ * std_grad_input.view(grad_output) 
                    + $ 4 \beta / (d(d-1))$ * cov_grad_input
        
        return grad_input
\end{lstlisting}
\end{algorithm*}

We quantify the computational overhead by measuring the average time required for one NVIDIA A100 GPU to execute both the forward and backward passes on the entire network for a batch size of 128 using the ImageNet dataset. These results are summarized in Table \ref{ta:time}. For the sake of comparison, we also include the latencies associated with adding Batch Normalization (BN) layers, revealing that our optimized VCReg implementation exhibits similar latencies to BN layers and is almost 5 times faster than the naive implementation.

% Put in appendix
\begin{table*}[h]
  \caption{\textbf{Average Time Required for One Forward and Backward Pass with Various Layers Inserted} Comparison of computational latencies across different configurations of ViT and ConvNeXt networks. The table demonstrates the efficacy of the optimized VCReg layer in terms of computational time, compared to both naive VCReg and Batch Normalization (BN) layers.}
  \label{ta:time}
  \vskip 0.1in
  \centering
  \scalebox{0.90}{
  \begin{tabular}{lcccccc}
    \toprule
    Network & Number of Inserted Layers & Identity & VCReg (Naive) & VCReg (Fast) & BN \\
    \midrule
    ViT-Base-32 & 12 & 0.223s & 1.427s & 0.245s & 0.247s  \\
    ConvNeXt-T & 18 & 0.442s & 2.951s & 0.471s & 0.468s  \\
    \bottomrule
  \end{tabular}}
\end{table*}

\section{Implementation Details}
\label{sec:appendix-detail}

\subsection{Transfer Learning Experiments with ImageNet Pretraining}

In conducting the transfer learning experiments, we adhered primarily to the training recipe specified by PyTorch \cite{NEURIPS2019_9015} for each respective architecture during the supervised pretraining phase. We abstained from pretraining any of the baseline models, instead opting to directly download the weights from PyTorch's own repository.
The only modifications applied were to the parameters associated with VCReg loss, and we experimented with $\alpha \in \{1.28, 0.64, 0.32, 0.16\}$ and $\beta \in \{0.16, 0.08, 0.04, 0.02, 0.01\}$.

For iNaturalist 18 \cite{van2018inaturalist} and Place205 \cite{zhou2014learning}, we relied on the experimental settings detailed in \cite{zbontar2021barlow} for the linear probe evaluation. 

Regarding Food-101 \cite{bossard2014food}, Stanford Cars \cite{krause20133d}, FGVC Aircraft \cite{maji13fine-grained}, Oxford-IIIT Pets \cite{parkhi2012cats}, Oxford 102 Flowers \cite{nilsback2008automated}, and the Describable Textures Dataset (DTD) \cite{cimpoi2014describing}, we complied with the evaluation protocol provided by \cite{chen2020simple, kornblith2021better}. An $L2$-regularized multinomial logistic regression classifier was trained on features extracted from the frozen pretrained network. Optimization of the softmax cross-entropy objective was conducted using L-BFGS, without the application of data augmentation. All images were resized to 224 pixels along the shorter side through bicubic resampling, followed by a 224 x 224 center crop. The $L2$-regularization parameter was selected from a range of 45 logarithmically spaced values between $0.00001$ and $100000$.

All experiments were run three times, with the average results presented in Table \ref{transfer_learning_1}.

\subsection{Transfer Learning Experiments with Kinetics pre-trained Models}
In conducting experiments with video-pretrained models, we utilize the publicly available code bases and model checkpoints provided for VideoMAE and VideoMAEv2 (\url{https://github.com/MCG-NJU/VideoMAE} and \url{https://github.com/OpenGVLab/VideoMAEv2}). For both VideoMAE and VideoMAEv2 we use ViT-Small and ViT-Base checkpoints. VideoMAE models are pre-trained on Kinetics-400 while VideoMAEv2 on Kinetics-710. We use the pre-trained checkpoint for ViViT-B (ViT-Base backbone) pre-trained on Kinetics-400 from HuggingFace. For evaluation, we adopt the inference protocol of 10 clips $\times$ 3 crops. For VCReg hyperparameters experiments with values for $\alpha \in {1, 3, 5}$ and $\beta\in\{0.1, 0..3, 0.5\}$. For the rest of the finetuning hyperparameters as well as the data pre-processing and evaluation protocol, we use the configuration for HMDB51 available in VideoMAE \cite{tong2022videomae} and its corresponding code base (linked above).

\subsection{Subclass Linear Probing Result with Network Pretrained on Superclass Label}

For our subclass linear probing experiments, we employed a ConvNeXt-Atto network. Each model was pretrained for 200 epochs using the superclasses, adhering to the same procedure detailed in the Appendix \ref{investigation}. Subsequent to this pretraining phase, we initiated a linear probing process using the subclass labels. This linear classifier was trained for 100 epochs, using a base learning rate of $0.016$ in conjunction with a cosine learning rate schedule. The optimizer used was AdamW, which worked to minimize cross-entropy loss with a weight decay set at $0.05$. We processed our training data in batches of 256.

\subsection{Long-Tail Learning Result}

For our long-tail learning experiments, we use ResNet-32 as a backbone for experiments on the CIFAR10-LT and CIFAR100-LT datasets.
We trained 100 epochs with batch size 256, Adam optimizer with two learning rate of $\{0.016, 0.008\}$ with a 10-epoch linear warm-up followed by a cosine annealing decay. The weight decay was set at $0.05$ and the norm layers were excluded from the weight decay. we experimented with $\alpha \in \{1.28, 0.64, 0.32, 0.16\}$ and $\beta \in \{0.16, 0.08, 0.04, 0.02, 0.01\}$.

\subsection{VCReg with Self-Supervised Learning Methods}

We train a ResNet-50 model in four different setups, using either the SimCLR loss or the VICReg loss with the ImageNet dataset.
The application of the VCReg is the same as described in Appendix \ref{investigation}.

We closely follow the original setting in \cite{chen2020simple} for SimCLR pretraining and \cite{bardes2021vicreg} for VICReg pretraining.

\textbf{Augmentation} For both methods, we use the same augmentation methods. Each augmented view is generated from a random set of augmentations of the same input image. 
We apply a series of standard augmentations for each view, including random cropping, resizing to 224x224, random horizontal flipping, random color-jittering, randomly converting to grayscale, and a random Gaussian blur. These augmentations are applied symmetrically on two branches \cite{geiping2022much}

\textbf{Architecture} 
For SimCLR, the encoder is a ResNet-50 network without the final classification layer followed by a projector. The projector is a two-layer MLP with input dimension 2048, hidden dimension 2048, and output dimension 256. The projector has ReLU between the two layers and batch normalization after every layer. This 256-dimensional embedding is fed to the infoNCE loss.

For VICReg, the online encoder is a ResNet-50 network without the final classification layer. The online projector is a two-layer MLP with input dimension 2048, hidden dimension 8192, and output dimension 8192.
The projector has ReLU between the two layers and batch normalization after every layer. This 8192-dimensional embedding is fed to the infoNCE loss.

For VCReg, we just applied the VCReg layers to the ResNet-50 network as described in the Appendix \ref{investigation}.

\textbf{Optimization} 
We follow the training protocol in \cite{zbontar2021barlow}.
For SimCLR experiments, we used a LARS optimizer and a base learning rate 0.3 with cosine learning rate decay schedule. We pretrain the model for 100 epochs with 5 epochs warm-up with batch size 4096.

For VICReg, we use a LARS optimizer and a base learning rate 0.2 using cosine learning rate decay schedule. We pretrain the model for 100 epochs with 5 epochs warm-up with batch size 4096.

\textbf{Evaluation} We follow the standard evaluation protocol as prescribed by \cite{misra2020self, zbontar2021barlow}, performing linear probing evaluations, on iNaturalist 18 \cite{van2018inaturalist} and Place205 \cite{zhou2014learning} datasets.

\section{Robustness to noise}\label{app:robustness-noise}
\begin{figure*}[ht]
\centering
\begin{subfigure}[VideoMAE-B]    
    {\includegraphics[width=0.45\textwidth]{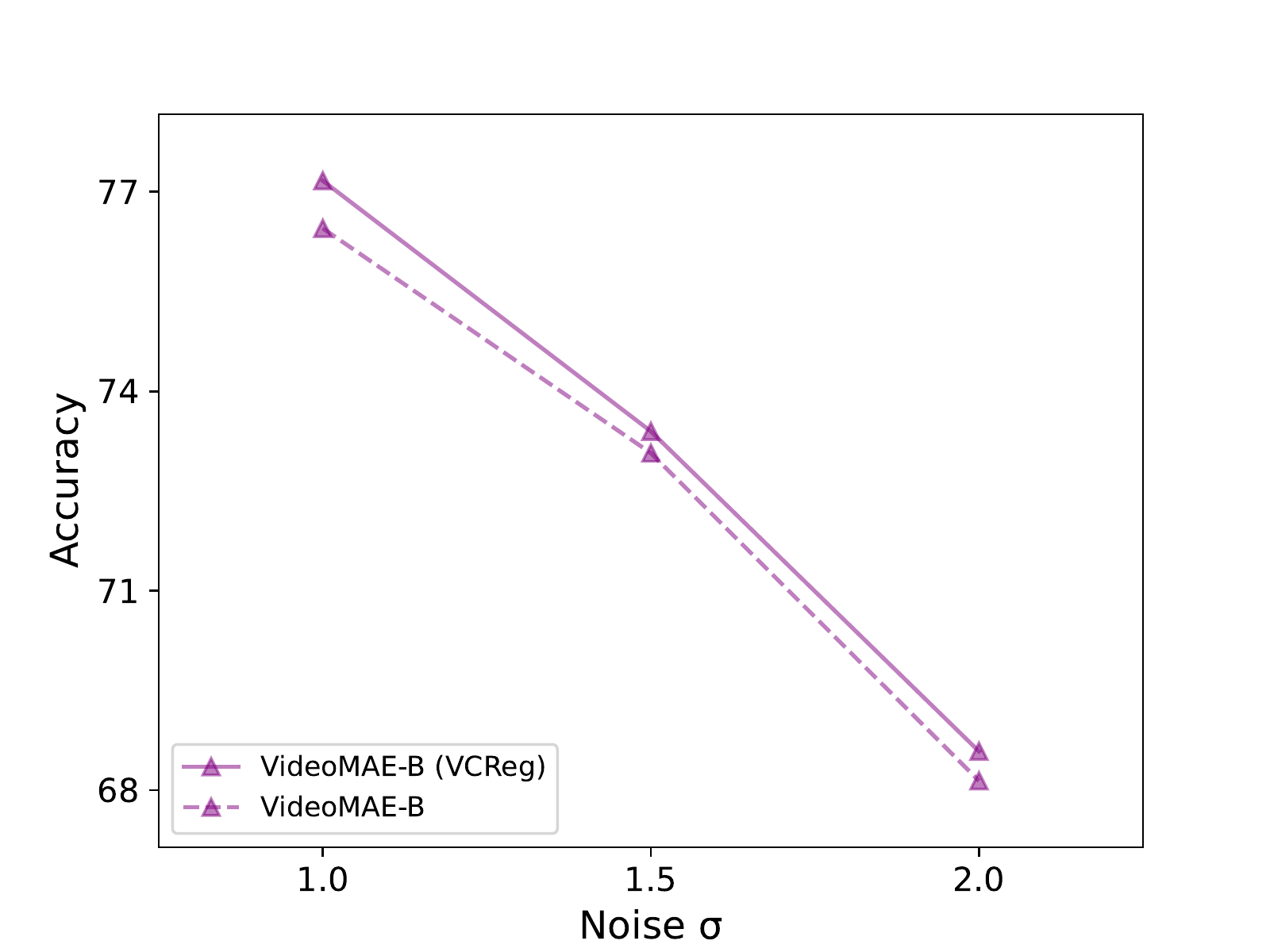}}
    % \caption{\textbf{Evaluating VCReg in the presence of noisy data} We report the top 1 accuracy of VideoMAE-B when fine-tuned on noisy versions of HMDB51. We corrupt the data with Gaussian noise with standard deviation $\sigma\in\{1, 1.5, 2\}$. VideoMAE-B (VCReg) outperforms its non-regularized version in this setting. }
     \quad
\end{subfigure}    
\begin{subfigure}[VideoMAEv2-B]
    {\includegraphics[width=0.45\textwidth]{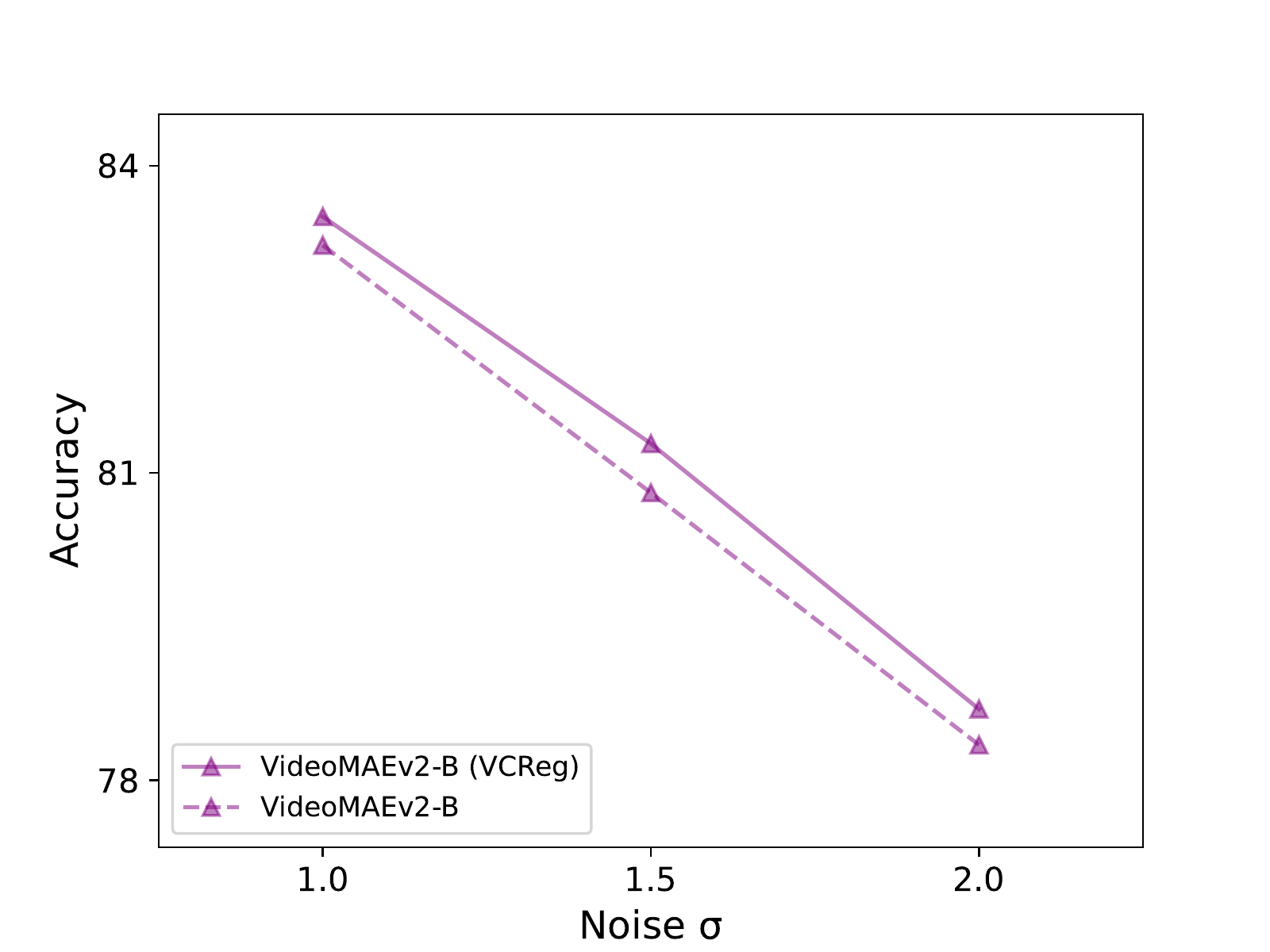}}
    % \caption{\textbf{Evaluating VCReg in the presence of noisy data} We report the top 1 accuracy of VideoMAEv2-B when fine-tuned on noisy versions of HMDB51. We corrupt the data with Gaussian noise with standard deviation $\sigma\in\{1, 1.5, 2\}$. VideoMAEv2-B (VCReg) outperforms its non-regularized version in this setting. }
\end{subfigure}
\caption{\textbf{Impact of VCReg amidst noisy data}: This figure shows the top-1 accuracy of VideoMAE-B and VideoMAEv2-B when fine-tuned for action recognition using HMDB51 with synthetic noise. We corrupt the data with Gaussian noise with standard deviation $\sigma\in\{1, 1.5, 2\}$. Models with VCReg outperform their non-regularized counterparts in this setting.}
 \label{fig:transfer-learning-noise-VideoMAEB}
\end{figure*}

This section provides additional results on measuring VCReg's ability to enhance transfer learning performance in the presence of noise. In these experiments we start with VideoMAE-B and VideoMAEv2-B networks (from section \ref{sec:transfer-with-video}) pre-trained on Kinetics-400 and Kinetics-710, respectively, then fine-tune them on HMDB51 corrupted with varying levels of Gaussian noise.  Figure \ref{fig:transfer-learning-noise-VideoMAEB} shows that VCReg models outperform their non-regularized counterparts in this setting.

\section{Two-Moon Dataset}
\label{tmsection}

\begin{figure}[t]
  \centering
  \includegraphics[height=67mm, width=120mm]{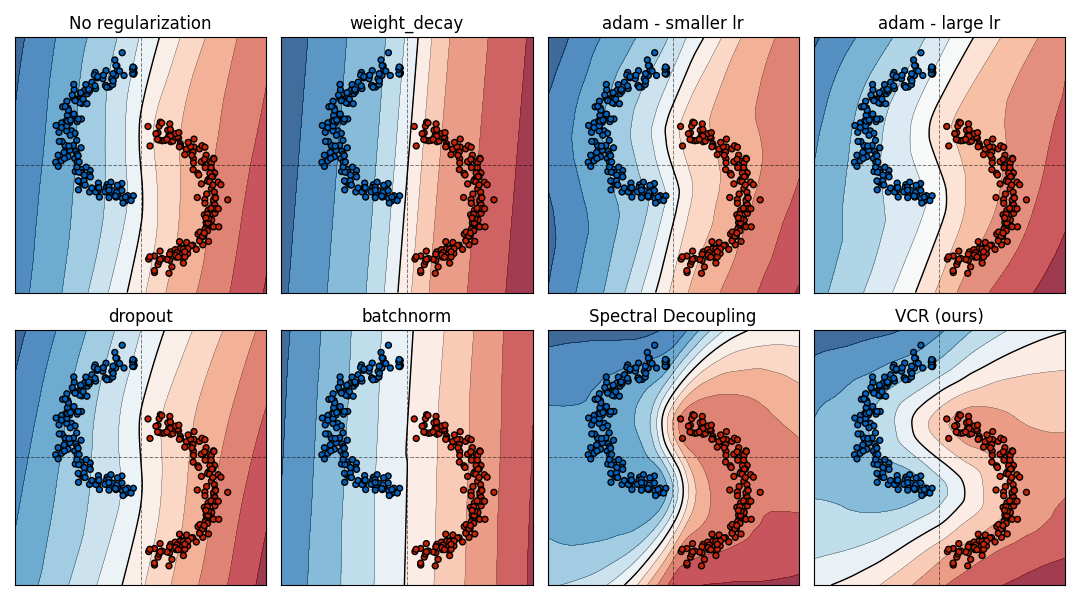}
  \caption{ \textbf{The effect of conventional regularization methods and the VCReg on a simple task of two-moon classification.} Shown decision boundaries are the average over 10 runs in which data points and the model initialization parameters are sampled randomly. Here, only the data points of one particular seed are plotted for visual clarity. It can be seen that conventional regularizations of deep learning seem not to help with learning a curved decision boundary.} 
  \label{two_moon}
\end{figure}

In alignment with the original gradient starvation study \cite{pezeshki2021gradient}, we notice that most regular routine regularization techniques do not sufficiently capture the necessary features for the ``two-moon'' dataset experiment. To evaluate our approach, we mirrored this setting and applied the VCReg during the training.

The synthetic ``two-moon'' dataset comprises two classes of points, each forming a moon-like shape. The gradient starvation study highlighted an issue where if the gap between the two moons is wide enough for a straight line to separate the two classes, the network stops learning additional features and focuses solely on a single feature. We duplicated this situation using a three-layer network and applied all the initially tested methods in the original study. The resulting decision boundary after training with the ``two-moon'' dataset is visualized in Figure \ref{two_moon}.

From the visualization,  it becomes apparent that not only does VCReg outperform other conventional regularization techniques in separation margins, but also it shows superior performance compared to spectral decoupling, a method specifically designed for this task. VCReg is effective in maximizing the variance while minimizing the covariance in the feature space, an achievement that is not obtained by other techniques such as L2, dropout \cite{hinton2012improving}, and batch normalization \cite{ioffe2015batch}. Consequently, these other techniques yield features that are less discriminative and informative.

\section{Miscellaneous}

\subsection{Compute Resources}

The majority of our experiments were run using AMD MI50 GPUs.
The longest pretraining for ConvNeXt-Tiny takes about 48 hours on 2 nodes, where each node has 8 MI50 GPUs attached.
We estimate that the total amount of compute resources used for all the experiments can be roughly approximated by $60 \text{ (days)} \times 24 \text{ (hours per day)} \times 8 \text{ (nodes)} \times 8 \text{ (GPUs per nodes)} = 92,160 \text{ (GPU hours)}$.

We are aware of potential environmental impact of consuming a lot of compute resources needed for this work, such as atmospheric $\text{CO}_2$ emissions due to the electricity used by the servers. However, we also believe that advancements in representation learning and transfer learning can potentially help mitigate these effects by reducing the need for data and compute resources in the future.

\subsection{Limitations} 

Due to a lack of compute resources, we were unable to conduct a large number of experiments with the goal of tuning hyperparameters and searching for the best configurations.
Therefore, the majority of hyperparameters and network configurations used in this work are the same as provided by PyTorch \cite{NEURIPS2019_9015}.
The only hyperparameters that were tuned were $\alpha$ and $\beta$, the coefficients for VCR.
All the other hyperparameters may not be optimal. 

In addition, all models were pretrained on the ImageNet \cite{deng2009imagenet} and \cite{krizhevsky2009learning} dataset, so their performances might differ if pretrained with other datasets containing different data distributions or different types of images (e.g., x-rays). 
We encourage further exploration in this direction for current and future self-supervised learning frameworks.

\section{Impact Statement}
This paper presents work whose goal is to advance the field of Machine Learning. There are many potential societal consequences of our work, none which we feel must be specifically highlighted here.

\end{document}